\documentclass[11pt,a4paper]{article}

\usepackage[utf8]{inputenc}
\usepackage[T1]{fontenc}
\usepackage{lmodern}
\usepackage{amsmath,amssymb}
\usepackage[ruled,vlined]{algorithm2e}
\usepackage{booktabs}
\usepackage{graphicx}
\usepackage{hyperref}
\usepackage[numbers]{natbib} 
\usepackage{url}         
\usepackage[margin=2cm]{geometry}
\usepackage{multirow}

\usepackage{caption}

\title{Prob-BBDM: a Probabilistic Brownian Bridge Diffusion Model for MRI sequence image-to-image translation}
\author{Martin Valls$^{1,4}$ \and Pascal Bourdon$^{1,4}$ \and Christine Fernandez$^{1,4}$ \and Guillaume Herpe$^{2,3}$ \and David Helbert$^{1,4}$}
\date{
$^1$ University of Poitiers, CNRS, XLIM, France\\
$^2$ University of Poitiers, CNRS, Laboratory of Applied Mathematics, France\\
$^3$ Poitiers University Hospital\\
$^4$ I3M common laboratory CNRS-Siemens Healthinners, Poitiers University Hospital and University of Poitiers, France
}

\begin{document}

\maketitle

\begin{abstract}
AI-driven image-to-image synthesis is rapidly advancing, with growing applications in medical imaging. Multi-modal image analysis plays a crucial role in optimizing examination quality, yet acquiring multiple imaging modalities in clinical settings remains resource-intensive and time-consuming, especially for 3D imaging. To address this challenge, we propose a novel image-to-image translation model based on Brownian Bridge Diffusion Models (BBDM), which synthesizes magnetic resonance imaging (MRI) sequences from 2D axial slices. Our approach integrates a variational encoder-guided diffusion mechanism, leveraging probabilistic image distributions to enhance synthesis quality. Evaluated on the BraTS 2021 dataset, our Probabilistic-BBDM (Prob-BBDM) achieves superior performance across multiple translation tasks, reaching up to $88.46\%$ SSIM and $26.09$ dB PSNR, with consistent improvements over baselines. Notably, our diffusion process requires only $4$ steps, making it computationally efficient while maintaining high-quality synthesis. To further validate generalizability, we test Prob-BBDM on an external third-party dataset, demonstrating consistent performance across domains. Additionally, we assess the clinical utility of the synthesized slices by using them as input to a pre-trained segmentation model. Tumor segmentation yields a Dice score of $88.71\%$ and an HD95 of $3.49$ mm, confirming that the synthesized slices preserve critical diagnostic information. These results highlight the potential of Prob-BBDM for high-quality, efficient, and generalizable MRI synthesis, offering a promising step toward improved medical image translation. \url{https://gitlab.xlim.fr/mvalls/Prob-BBDM}
\end{abstract}

\section{Introduction}\label{sec:introduction}

Achieving high-quality clinical MRI examinations is challenging, requiring a balance between acquisition time and the minimum necessary sequences for diagnosis, as well as between image quality and signal-to-noise ratio. These challenges are further compounded by the increasing demand for MRI. To address this, deep learning-based solutions have been developed to reduce acquisition time without compromising image quality or diagnostic performance. Some commercially available solutions claim up to an 80\% reduction in acquisition time (\textit{e.g.}, Swift MR \citep{airs_swiftmr_2025}, SubtleMR \citep{subtle_medical_2019}. \\
With the rise of generative AI, algorithms have demonstrated efficiency in generating synthetic medical images\citep{dayarathna_deep_2024, wang_diffusion_2025, guo_towards_2025}. The objective of our study is to evaluate whether it is feasible to generate additional contrast-weighted images from a single acquired MRI sequence and to determine the most suitable model for this translation task. Image translation aims at taking an image from a source domain and obtaining a matched image from a target domain. Deep learning algorithms have succeeded in creating a mapping between the distributions of these images through various processes \citep{cao_survey_2023, kazerouni_diffusion_2023}.
Generative models such as Generative Adversarial Networks (GAN) \citep{armanious_medgan_2019, kong_breaking_nodate}, Transformers \citep{liu_one_2023, pan_2d_2023} and Diffusion Models  (DM) \citep{linguraru_cross_conditioned_2024, atli_i2i-mamba_2025}, have considerably improved the quality and versatility of image-to-image translation. While 3D diffusion models  \citep{kim_adaptive_2024} have shown promising results, they remain computationally expensive and memory intensive, often requiring trade-offs in resolution and training stability \citep{khader_medical_2023}. Consequently, most diffusion-based medical image translation methods operate on 2D slices \citep{goswami_learndiff_2025,  ozbey_unsupervised_2023}. In this work, we adopt a 2D formulation to enable high-resolution modeling and efficient training. \\
GANs \citep{goodfellow_generative_2014} demonstrated high quality in the images generated \citep{wang_cyclesgan_2024}, and those with fast sampling times. On the other hand, this type of network has been shown to have a weakness: mode collapse, which results in a loss of diversity in the images synthesized \citep{dhariwal_diffusion_2021, xiao_tackling_2022}.
Competing with GAN, Transformers \citep{shamshad_transformers_2023, vaswani_attention_2023} uses an attention-based mechanism for efficient feature learning, offering better image adaptation than traditional convolutional neural networks (CNN) \citep{oshea_introduction_2015}. Vision Transformer (ViT) \citep{dosovitskiy_image_2021} processes images as patch sequences, making it suitable for large datasets but resource-intensive.
Last but not least, the DM \citep{dickstein_deep_2015, zhu_2024} were able to generate images of great quality and diversity \citep{kebaili_multi-modal_2025}. Through a Markov chain, these probabilistic models create a link between the source domain and a Gaussian noise domain. This link is then reversed, by coding the target domain in order to recover the source domain image during translation.

In this context, BBDM \citep{li_bbdm_2023} models were developed to establish a theoretical link between diffusion domains through a stochastic process. In medical imaging translation tasks, precision and robustness are critical for clinical diagnosis and treatment monitoring \citep{dayarathna_deep_2024} as the goal is to synthesize reliable images. To address this challenge, our key contributions are summarized as follows: 
\begin{itemize}
    \item A probabilistic Brownian Bridge Diffusion framework with prior-posterior latent alignment for anatomically consistent MRI sequence translation.
    \item We demonstrate that high-quality MRI synthesis can be achieved with as few as $4$ diffusion steps, substantially reducing computational cost compared to standard diffusion approaches.
    \item Validation on BraTS 2021 \cite{mohan_rsna-asnr-miccai_2021} and an external clinical dataset, including downstream tumor segmentation analysis confirming preservation of clinically relevant structures.
\end{itemize}
This work contributes to advancing multi-modal medical image synthesis with a computationally efficient and clinically relevant approach.

\begin{figure*}
    \centering
    \includegraphics[width=0.9\textwidth]{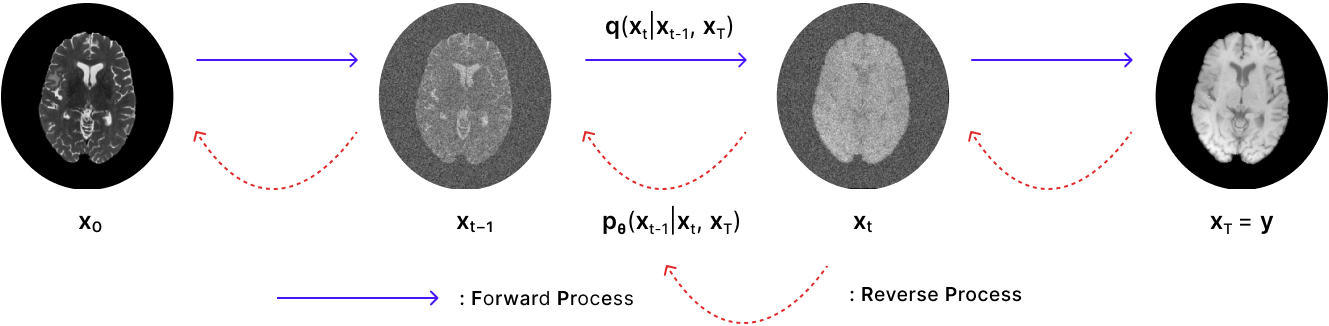}
    \caption{The graphical model of the Brownian Bridge Diffusion Model. Originally, the denoising diffusion probabilistic model (DDPM) uses the target image as the starting point $\mathbf{x}_0$ and pure Gaussian noise as the end point $\mathbf{x}_T = \mathbf{y}$ of the forward diffusion process. The BBDM differs by terminating the diffusion process towards the source image with additional noise. This method creates a bridge between the target image and the source image.}
    \label{fig:graphical model}
\end{figure*}

\section{Related Work}

\subsection{Diffusion Models}
Recent progress in image generation using deep learning is due in particular to advanced diffusion models. Introduced by Ho \textit{et al.} \citep{ho_denoising_2020}, these models can be broken down into two phases, a forward diffusion process and a reverse diffusion process. Starting with data $\bold{x}_0 \sim q(\bold{x}_0)$, the forward process create a Markov chain such that we can formulate the distribution of states $\bold{x}_1,...,\bold{x}_T$ given $\bold{x}_0$ by: 

\begin{equation}
    q(\bold{x}_1,...,\bold{x}_T | \bold{x}_0) = \prod^{T}_{t=1} q(\bold{x}_t | \bold{x}_{t-1}).
\end{equation}
with $ q(\bold{x}_t | \bold{x}_{t-1}) = \mathcal{N}(\bold{x}_t; \sqrt{1-\beta_t}\bold{x}_{t-1},  \beta_t \bold{I})$ a normal distribution with a variance schedule  $\beta_1,...,\beta_T$. After T operations, the data $\bold{x}_T$ has become a Gaussian noise and  $p(\bold{x}_T) = \mathcal{N}(0,\bold{I})$.
At any timestep $t$, $\bold{x}_t$ can be formulated as:

\begin{align}
    \bold{x}_t &= \sqrt{1-\beta_t} \bold{x}_{t-1} + \sqrt{\beta_t} \mathbf{\epsilon}_{t-1}, \\
    &= \sqrt{1 - \tilde{\beta}_t} \bold{x}_{0} + \sqrt{\tilde{\beta}_t} \mathbf{\epsilon}_t. 
\end{align} 
with $\tilde{\beta}_t = \prod_{i=1}^{t} \beta_t$ and $\mathbf{\epsilon}_1, ... , \mathbf{\epsilon}_T$ are independent and identically distributed (i.i.d.) samples from the standard multivariate normal distribution $\mathcal{N}(\bold{0},\bold{I})$. \\

The forward diffusion process can also be formulate in continuous time in a stochastic differential equation form:

\begin{equation}
    d\bold{x}_t = - \frac{1}{2}\beta(t)\bold{x}_t dt + \sqrt{\beta(t)}dW_t.
\end{equation}
where $W_t$ is a Wiener Process also called Brownian motion.

On the other hand the reverse process try to approximate the posterior distribution:

\begin{equation}
p(\bold{x}_0,...,\bold{x}_{T-1} \mid \bold{x}_T) = \prod^{T}_{t=1} p_\theta(\bold{x}_{t-1} \mid \bold{x}_t).
\end{equation}

We define \( p_\theta(\bold{x}_{t-1} \mid \bold{x}_t) = \mathcal{N}(\bold{x}_{t-1};\mu_\theta(\bold{x}_0,t), \delta_\theta(\bold{x}_0,t)) \), where \( \theta \) are the weights of the diffusion model.

\subsection{Brownian Bridge Diffusion}
A Brownian bridge is a continuous-time stochastic model in which the probability distribution during the diffusion process is conditioned by the beginning point $\bold{x}_{0}$ and ending point $\bold{x}_{T} = \bold{y}$. Let ($\bold{x}, \bold{x}_T$) be a matched pair of data as target and source sequences. We can define at each time step $t$, starting at $\bold{x}_0 \sim p(\bold{x}_0)$ from target sequence domain, the forward diffusion process to:

\begin{align}
\bold{x}_t &= (1 - m_t)\bold{x}_0 + m_t\bold{x}_T + \sqrt{\delta_t}\mathbf{\epsilon}_t, \\
q(\bold{x}_t |\bold{x}_0,\bold{x}_T) &= \mathcal{N}(\bold{x}_t; (1 - m_t)\bold{x}_0 + m_t\bold{x}_T, \delta_t \mathbf{I}).
\end{align}
where $m_t = t/T$, $\delta_t =\frac{t(T - t)}{T}$ follow a variance scheduler for the Brownian Bridge diffusion process. \\

The reverse diffusion process, which denoises from the source sequence $\bold{x}_T$ to the initial target state $\bold{x}_0$, is defined as:

\begin{equation}
  p_\theta(\bold{x}_{t-1} \mid \bold{x}_t, \bold{x}_T) = \mathcal{N}\!\left(\bold{x}_{t-1};\, \mu_\theta(\bold{x}_t, t),\, \tilde{\delta}_t \bold{I}\right).
\end{equation}

The conditional mean $\mu_\theta(\bold{x}_t, t)$ is predicted by the diffusion U-Net with parameters $\theta$ and corresponds to the predicted mean of the denoised state. The variance term $\tilde{\delta}_t$ is not learned but is analytically computed from the forward process variances $\delta_t$. This Brownian bridge formulation enables a smooth transition between the initial and final states, making it suitable for image-to-image translation from a source domain to a target domain. In standard DDPMs \citep{ho_denoising_2020}, the forward process ends in pure Gaussian noise, whereas in BBDM the conventional conditional input $\bold{y} = \bold{x}_T$ is designated as the final forward state.
Following DDPM, training is performed by minimizing the Evidence Lower Bound (ELBO):

\begin{equation}
\begin{split}
\mathbb{E}_q \Bigg[ 
&\underbrace{D_{\mathrm{KL}}(q(\bold{x}_T \vert \bold{x}_0,\bold{x}_T) \,\Vert\, p_\theta(\bold{x}_T\vert \bold{x}_T))}_{L_T}
 - \underbrace{\log p_\theta(\bold{x}_0\vert \bold{x}_1,\bold{x}_T)}_{L_0} \\
&\quad+ \sum_{t=2}^T 
\underbrace{D_{\mathrm{KL}}\!\left(q(\bold{x}_{t-1}\vert \bold{x}_t,\bold{x}_0,\bold{x}_T) \,\Vert\, p_\theta(\bold{x}_{t-1}\vert \bold{x}_t,\bold{x}_T)\right)}_{L_{\theta}}
\Bigg].
\end{split}
\end{equation}

\begin{figure*}
    \centering
    \includegraphics[width=0.7\textwidth]{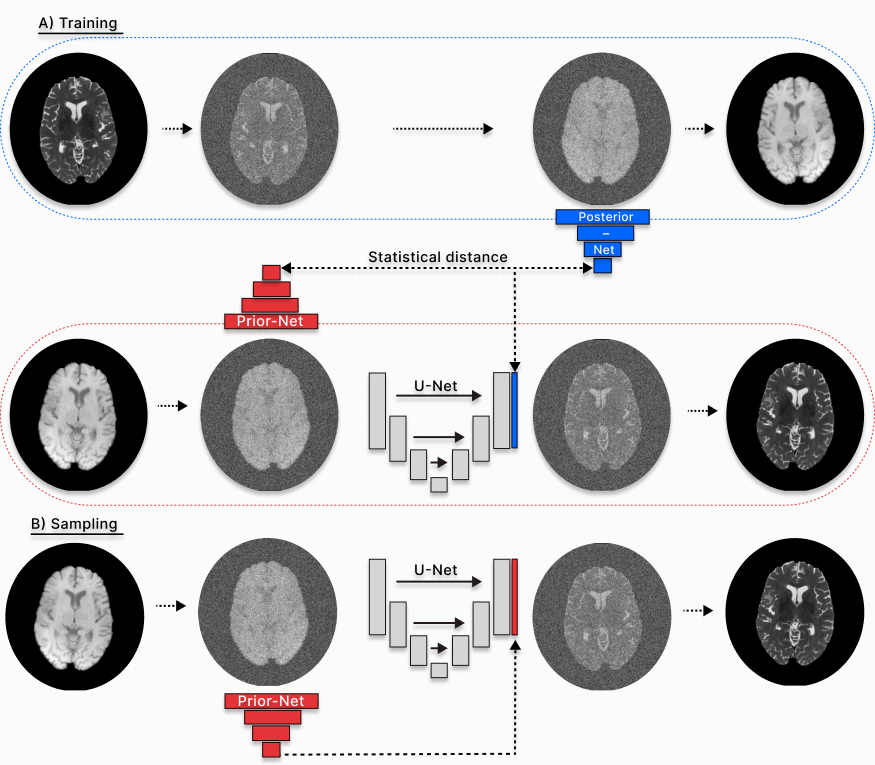}
    \caption{Overview of the Prob-BBDM framework for MRI sequences translation.
(A) Training phase: A target sequence is progressively noised by iteratively interpolating it with the source sequence and adding Gaussian noise. Posterior-Net learned the distribution of the $\mathbf{x}_t$. The Prior-Net conditions on the source image to generate a prior distribution over intermediate noisy states, which guides the U-Net during the denoising process. The Prior-Net is optimized by minimizing the statistical distance between the learned posterior and the prior.
(B) Sampling phase: During inference, the learned Prior-Net provides the initial latent distribution, and the U-Net progressively transforms this distribution into a target modality image. The full process allows efficient source-to-target translation with high fidelity and low computational cost.}
    \label{fig:Training and sampling}
\end{figure*}

During training, $L_T$ is constant since $\bold{x}_T$ is the source, and $L_0$ is handled by a discrete decoder as in \citep{ho_denoising_2020}. The main objective is therefore to approximate  
$q(\bold{x}_{t-1}\!\mid\!\bold{x}_t,\bold{x}_0,\bold{x}_T)$ with  
$p_\theta(\bold{x}_{t-1}\!\mid\!\bold{x}_t,\bold{x}_T)$.

Because $\bold{x}_{t-1}$ is unknown in the forward process, the posterior is rewritten via Bayes’ theorem using the Gaussian transitions of the forward process:
\begin{equation}
\begin{split}
q(\bold{x}_{t-1}\vert \bold{x}_t,\bold{x}_0,\bold{x}_T)
&= 
q(\bold{x}_t \vert \bold{x}_{t-1},\bold{x}_T)
\frac{q(\bold{x}_{t-1} \vert \bold{x}_0,\bold{x}_T)}{q(\bold{x}_t \vert \bold{x}_0,\bold{x}_T)} \\
&= \mathcal{N}\!\left(\bold{x}_{t-1};\, \tilde{\mu}_\theta(\bold{x}_t,\bold{x}_T),\, \tilde{\delta}_t \bold{I}\right).
\end{split}
\end{equation}

Since the product of Gaussian densities remains Gaussian, the mean can be written in linear form:
\begin{equation}
\begin{split}
\tilde{\mu}_\theta(\bold{x}_t,\bold{y}) 
&= c_{x_t}(t)\,\bold{x}_t 
  + c_{y}(t)\,\bold{y} \\
&\quad + c_{\epsilon}(t)\!\left(m_t(\bold{y}-\bold{x}_0) 
  + \sqrt{\delta_t}\,\mathbf{\epsilon}_t\right) \\
&= c_{x_t}(t)\,\bold{x}_t 
  + c_{y}(t)\,\bold{y} \\
&\quad + c_{\epsilon}(t)\,\mathbf{\epsilon}_\theta(\bold{x}_t,t),
\end{split}
\end{equation}
where the coefficients $c_{x_t}(t)$, $c_{y}(t)$ and $c_{\epsilon}(t)$ are determined analytically from the forward process Eq. (6-7), and $\mathbf{\epsilon}_\theta(\bold{x}_t,t)$ denotes the noise estimated by the denoising network.

\medskip
To simplify optimization, l'ELBO se résume à l'objectif de prédiction du bruit suivant :

\begin{equation}
L_\theta 
= \mathbb{E}_{t,\bold{x}_0,\bold{x}_T,\mathbf{\epsilon}_t}[
  c_{\epsilon}(t)\,
  \big\|
    \tfrac{t}{T}(\bold{x}_T-\bold{x}_0)
    + \sqrt{\delta_t}\,\mathbf{\epsilon}_t
    - \mathbf{\epsilon}_\theta(\bold{x}_t,t)
  \big\|^2],
\end{equation}
where $\mathbf{\epsilon}_t\sim\mathcal{N}(0,\mathbf{I})$ is the noise injected in the forward process at step $t$.  
Unlike conditional DDPMs \citep{choi_ilvr_2021}, the network does not receive $\bold{x}_T$ as an explicit conditioning input during the reverse process, but instead relies on the Brownian bridge formulation, where $\bold{x}_T$ acts as a forward diffusion endpoint. This design avoids overly restrictive conditioning at each denoising step while still constraining the overall diffusion trajectory.
Due to the Brownian bridge constraint between known endpoints, the reverse process operates within a restricted solution manifold rather than the full image space explored by standard DDPMs. This structural constraint reduces the effective entropy of the generative process, which theoretically lowers the number of sampling steps required to recover the target image.

\section{Theory and Methods}

\subsection{Prob-BBDM}
To enhance the model’s ability to synthesize the target image reliably, we introduce a Probabilistic Brownian Bridge Model. This approach restricts the model’s freedom to generate images that deviate excessively from the target. Equation (9) defines the objective as to determine the noise component, referred to as the Brownian diffusion objective. This formulation accounts for both the initial state and the desired target while allowing controlled stochastic exploration of the latent space. To achieve this, we integrate a probabilistic framework and modify the architecture of standard U-Net diffusion models.

\subsubsection{Forward Process}
Based on the work of Kohl \textit{et al}. \citep{kohl_probabilistic_2019}, a couple (prior, posterior) of conditional variational encoder \citep{sohn_learning_nodate} is used to learn how to model the distribution of our data $\bold{x}_t$ in a latent space as Gaussian of parameters $(\mu(\bold{x}_t); \sigma(\bold{x}_t))$.

\begin{align}
\bold{x}_t &=  \bold{x}_0 + \frac{t}{T} (\bold{x}_T - \bold{x}_0) + \sqrt{\delta_t}\mathbf{\epsilon_t}, \\
q(\bold{x}_t |\bold{x}_0, \bold{x}_T) 
&= \mathcal{N}(\mu_{post}(\bold{x}_t, \rho); \sigma_{post}(\bold{x}_t, \rho) \mathbf{I}).
\end{align}
where $\rho$ is the weight of the posterior encoder net which estimates the distribution parameters at time step t.

\subsubsection{Reverse Process}
During training, the reverse diffusion steps generates $\bold{x}_{t-1}$:
\begin{align}
    \bold{x}_{t-1} &= \bold{x}_t - \mathbf{\epsilon}_\theta(\bold{x}_t, \bold{x}_T, t), \\
    p_\theta(\bold{x}_{t-1}|\bold{x}_t, \bold{x}_T) &= \mathcal{N}(\mu_{prior}(\bold{x}_t, \omega); \sigma_{prior}(\bold{x}_t, \omega) \mathbf{I}).
\end{align}
 $\bold{x}_{t-1}$ est la prédiction déduite du modèle de diffusion et $\omega$ le poids du réseau d'encodage a priori.
 
\subsubsection{Training Objective}
The training objective combines the Brownian bridge diffusion loss with a probabilistic regularization term that aligns the prior and posterior latent distributions.
The two components from forward and reverse process allow us to compute the Kullback–Leibler divergence $D_{KL}(P_\theta||Q) = \mathbb{E}_{\bold{x} \sim P_\theta} [\log P_\theta(\bold{x})] - \mathbb{E}_{\bold{x} \sim P_\theta} [\log Q(\bold{x})]$ between distributions. 
Specifically, the Posterior-Net and Prior-Net parameterize Gaussian distributions over intermediate noisy states at each diffusion step. 

During training, the KL divergence is computed between the posterior distribution conditioned on both source and target images and the prior distribution conditioned only on the source image:

\begin{equation}
    L_{\rho, \omega} = D_{KL}(p_\theta(\bold{x}_{t-1}|\bold{x}_t, \bold{x}_T) || q(\bold{x}_t|\bold{x}_0, \bold{x}_T)).
\end{equation}

In other words, at each iteration, the model calculates the statistical distance to the desired translated target. This transition enables the model to maintain its probabilistic setting even when faced with unknown data. These additions increase the reliability of our synthesized image only at the cost of a very slight increase in computation time. By penalizing the model when distributions are too far apart, the model is not only focusing on denoising but also to get closer to the target. 

While standard BBDM conditions the diffusion process implicitly through the Brownian bridge endpoints $\mathbf{x}_T$, our approach additionally injects the source image as conditioning input to the denoising network $\mathbf{\epsilon}_\theta$, improving anatomical guidance at each diffusion step.

From Eq. (13), the residual between the source and target endpoints can be expressed as:

\begin{equation}
   \bold{x}_T - \bold{x}_0 = \frac{T}{t} (\bold{x}_t - \bold{x}_0 - \sqrt{\delta_t}\mathbf{\epsilon_t}).
\end{equation}

Since $\bold{x}_t - \bold{x}_0$ correspond to all the noise added during $t-1$ first steps, our objective is to detect the noise $\mathbf{\epsilon}_\theta(\bold{x}_t, \bold{x}_T, t)$  as the difference between our source $\bold{x}_T$ and the target $\bold{x}_0$. The exploration of space is then limited, and the aim becomes to reduce the difference to get closer to the target representation.

The overall objective $\mathcal{L}_\Theta$ can therefore be defined as a combination of our diffusion
 objective $ L_\theta$ and the source-target matching by statistical distance reduction with $L_{\rho, \omega}$.
 \begin{equation}
\begin{split}  
    \mathcal{L}_\Theta  &= L_\theta + \lambda  L_{\rho, \omega}, \\
    &= \underbrace{\mathbb{E}_{t, \bold{x}_0, \bold{x}_T}[||(\bold{x}_T - \bold{x}_0) - \mathbf{\epsilon}_\theta(\bold{x}_t, \bold{x}_T,  t) ||^2]}_{L_\theta}  \\
    & ~~~~~~~~~+ \lambda  \underbrace{D_{KL}(P_\theta||Q)}_{L_{\rho, \omega}}.
\end{split}
\end{equation}
The weighting factor $\lambda$ controls the relative contribution of the probabilistic regularization and is fixed empirically based on validation performance. 

\begin{algorithm}
\caption{Training of Probabilistic Brownian Bridge Model}
\label{alg:training}
\KwIn{Sequence pair $(\bold{x}_0, \bold{x}_T)$ as target and source}
\KwData{Max time step $T$, noise schedule $\delta_t$, $m_t$, predicted noise $\mathbf{\epsilon}_\theta$, predicted state parameters $(\mu_{prior}, \sigma_{prior})$, predicted target parameters $(\mu_{post}, \sigma_{post})$, weighting factor $\lambda$}
\While{not converged}{
    Sample $(\bold{x}_0, \bold{x}_T) \sim q(\bold{x}_0) \times q(\bold{x}_T)$\; 
    Sample $t \sim \text{Uniform}(1, \dots, T)$\; 
    Sample $\mathbf{\epsilon} \sim \mathcal{N}(0, \mathbf{I})$\;
    \textbf{Forward diffusion:} \\
    \Indp $\bold{x}_t = (1 - m_t)\bold{x}_0 + m_t \bold{x}_T + \sqrt{\delta_t} \mathbf{\epsilon}$\;
    \Indm
    \textbf{Posterior encoding:} \\
    \Indp $\bold{x}_t \sim \mathcal{N}(\mu_{post}, \sigma_{post})$\;
    \Indm
    \textbf{Prior encoding:} \\
    \Indp $\bold{x}_{t-1} \sim \mathcal{N}(\mu_{prior}, \sigma_{prior})$\;
    \Indm
    \textbf{Gradient step:} \\
    \Indp
    $\nabla_{\Theta} \left\| L_{\theta}(\bold{x}_T - \bold{x}_0, \mathbf{\epsilon}_\theta(\bold{x}_t, \bold{x}_T,  t)) + \lambda L_{\rho, \omega}(\bold{x}_{t-1},\bold{x}_t) \right\|$\;
    \Indm
}
\end{algorithm}
\begin{algorithm}
\caption{Sampling from the Probabilistic Brownian Bridge Model}
\label{alg:sampling}
\KwIn{Conditional sequence $\bold{x}_T$ as source}

\While{$t > 0$}{
    Sample $\mathbf{\epsilon}_t \sim \mathcal{N}(0, \mathbf{I})$\;

    \textbf{Prior encoding:} \\
    \Indp $\bold{x}_{t} \sim \mathcal{N}(\mu_{prior}, \sigma_{prior})$\;
    \Indm

    \textbf{Prediction of } $\hat{\bold{x}}_0$: \\
    \Indp $\hat{\bold{x}}_0 = \bold{x}_T - \mathbf{\epsilon}_\theta(\bold{x}_t, \bold{x}_T, t)$\;
    \Indm

    \textbf{Sampling step:} \\
    \Indp $\bold{x}_{t-1} = c_{\bold{x}t}\hat{\bold{x}}_0 + c_{\bold{x}_Tt}\bold{x}_T + c_{\mathbf{\epsilon}t}\mathbf{\epsilon}_\theta(\bold{x}_t, \bold{x}_T, t) + \sqrt{\tilde{\delta}} \mathbf{\epsilon}_t$\;
    \Indm
}
\Return{$\hat{\bold{x}}_0$ as generated target}
\end{algorithm}

\subsubsection{Implementation details} \label{Implementation details}
As illustrated in \autoref{fig:Training and sampling}, Prob-BBDM combines a Brownian Bridge Diffusion Model with a probabilistic latent conditioning strategy. In addition to the diffusion U-Net, the framework includes two lightweight convolutional encoder networks (Prior-Net and Posterior-Net) that model Gaussian distributions over intermediate noisy states.
Both Prior-Net and Posterior-Net are fully convolutional encoders composed of four resolution levels with 32, 64, 128, and 192 feature channels, and three convolutional layers per level.
During training (\autoref{fig:Training and sampling}A), the Posterior-Net learns the distribution of intermediate noisy states. The Prior-Net conditions on the source image to model a prior distribution over these states. The network is optimized by minimizing the Kullback–Leibler divergence between the posterior and prior distributions.
The latent variables sampled from these distributions are injected into the diffusion model via a feature-combination module, implemented as a stack of four convolutional layers. This module fuses the sampled latent representation with intermediate diffusion features to modulate the generation process through noise detection.
During inference (\autoref{fig:Training and sampling}B), the Posterior-Net is discarded and the learned Prior-Net alone provides the latent conditioning, guiding the U-Net to efficiently generate the target modality.
The diffusion backbone is a 2D U-Net, with six input channels, by concatenating the noisy intermediate state and the source image. It employs residual blocks, multi-head self-attention at resolutions 32, 16, and 8, and scale-shift normalization. The Prior-Net, Posterior-Net, and feature-combination module are implemented as auxiliary components within the same diffusion model, but are distinct from the core U-Net backbone and are used exclusively for probabilistic conditioning.

\section{Experiments}
\subsection{Dataset}
Our model was trained on the BraTS 2021 dataset consisting of 1250 images of size 240×240×155. Initially created for a tumor segmentation challenge, this dataset includes four MRI sequences (T1, T1ce, T2, and FLAIR) along with a pathology mask. For these experiments, we have set aside the T1 contrast-enhanced sequence and the mask to focus on translation between the base sequences. Pre-processing includes extraction of mid axial slices to ensure complex structure and the presence of a pathological zone. The 2D sections were then center cropped to 192×192 to reduce memory costs with a split (training, validation, test) of (1000, 200, 50) subjects.
During training, standard data augmentation techniques were applied, including random flip and small intensity variations, to improve generalization. The hyper-parameters were selected based on performance on the validation set and remained fixed throughout the experiments. The dataset split was fixed throughout all experiments to ensure reproducibility and consistent comparison across methods.

\subsection{Competing Methods}
BBDM \citep{li_bbdm_2023} Brownian Bridge Diffusion Model, which replaces traditional conditioning with a diffusion bridge formulation. This model serves as a foundational diffusion-based alternative, learning the source-to-target translation via a stochastic diffusion process without variational conditioning.\\
MaskGAN: \citep{phan_structure_preserving_2023} an advanced GAN variant specifically designed for medical imaging, which introduces an auxiliary mask generation branch. This branch focuses on explicit anatomical structure preservation, encouraging the network to consistently reconstruct critical regions such as tissues and pathology boundaries.\\
Pix2Pix \citep{pix2pix2017}: a conditional Generative Adversarial Network (cGAN) that incorporates source images as conditioning inputs to the generator. The model learns a direct mapping from source to target domain using an adversarial loss combined with an L1 reconstruction loss to encourage structural fidelity.\\
ResViT \citep{dalmaz_resvit_2022}: a hybrid model that integrates Vision Transformers with residual convolutional blocks. The convolutional components retain local spatial detail, while the transformer layers capture long-range dependencies, making it particularly suitable for translation tasks requiring context-awareness and precise localization.\\
SelfRDB \citep{arslan_self-consistent_2025}: a diffusion-based method that extends Brownian bridge formulations through self-consistent recursive conditioning. It captures a soft prior on the source modality in order to achieve greater robustness for noise measurement, thereby improving fidelity and stability in medical image translation.\\
To ensure a fair and consistent comparison, all competing methods were implemented using the official code released by the original authors when available. We follow the configurations and training protocols reported in the corresponding publications. All models were trained on the same preprocessed dataset using identical data splits, input resolution, and evaluation metrics. Hyper-parameters such as learning rate, batch size, optimizer, and number of training epochs were set according to the authors' recommendations for medical image translation tasks.

\subsection{Modeling Procedures}
We implemented using PyTorch and for optimization we utilized the Adam optimizer with an initial learning rate of 1.e-4 coupled with a scheduler that reduces the learning rate upon reaching a plateau, and a batch size of 4. Training was performed on one NVIDIA A-100 40GB GPU. Our model converged within 100 epochs, achieving stable performance on both the training and validation sets. We evaluate our framework on three distinct and challenging types of image-to-image translation consisting of T1-to-T2, T2-to-Flair, and Flair-to-T1 sequence synthesis. Inference times are listed in \autoref{tbl:SamplingStepStudy}. To evaluate the performance of our model, we used the Structural Similarity Index Measure (SSIM) to assess the structural similarity between the 2D synthesized image and the target. This method quantifies the quality of both structural and perceptual information in the translated image relative to the target. We also use Peak Signal to Noise Ratio (PSNR), which calculates the loss of information after reconstruction of the image that has passed through the model and been compressed. Since our algorithm performs generation rather than reconstruction, the use of PSNR could be questionable. It does however provide a pixel-level comparison metric between different models. For segmentation, quantitative metrics such as Dice Similarity Coefficient (DSC) and Hausdorff Distance (95\%) (HD95) were computed across the different methods. The DSC measures the overlap between predicted and ground-truth segmentations, with a higher score indicating better agreement, while the HD evaluates the maximum surface distance between segmented boundaries, where lower values reflect more accurate boundary alignment. To statistically validate our results, we used the Friedman test, a non-parametric test for multiple related samples that does not assume a normal distribution of measurements. When significant differences were detected, post hoc pairwise comparisons were performed using the Wilcoxon test with Holm correction to control for multiple comparisons and reduce the risk of type I errors.

\section{Results}
\begin{figure*}
    \centering
    \includegraphics[width=\textwidth]{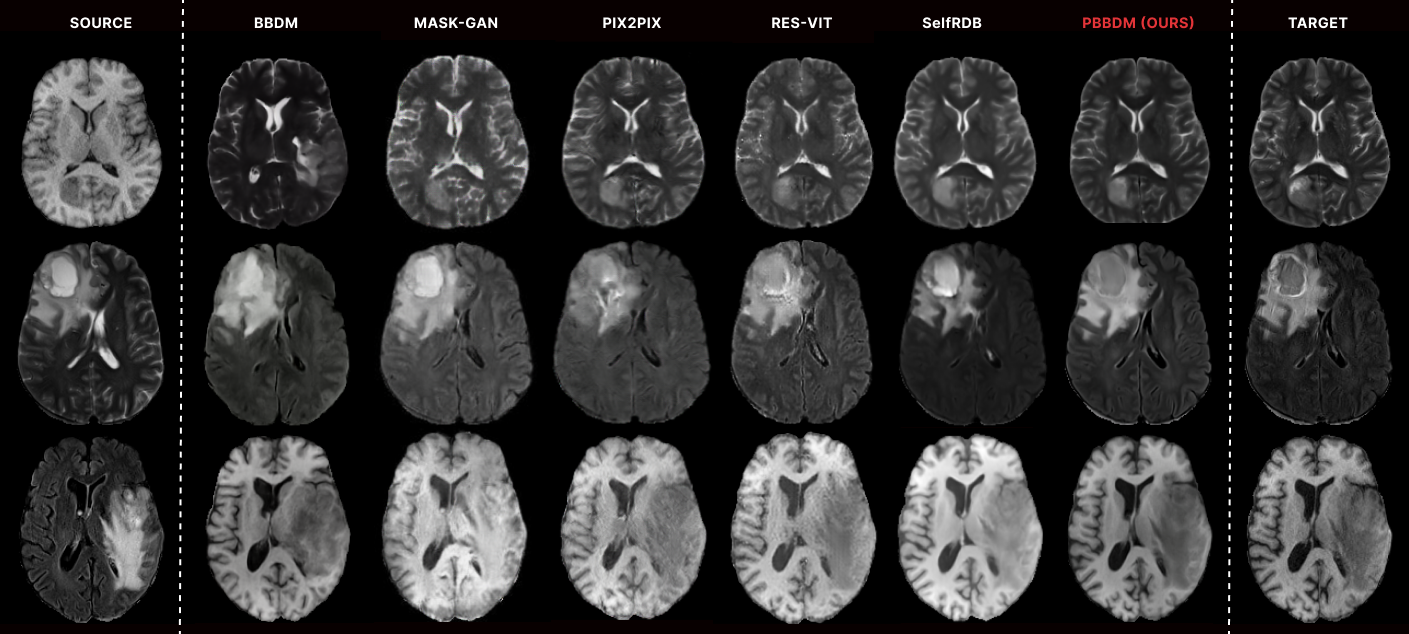}
    \caption{Qualitative comparison of synthesized MRI sequences from source images using reference models. Each row represents a different modality translation task: (1st row) T1-to-T2, (2nd row) T2-to-FLAIR, and (3rd row) FLAIR-to-T1. From left to right: source image, outputs from Pix2Pix, MaskGAN, ResViT, BBDM, SelfRDB, our proposed Prob-BBDM, and the ground-truth target image. The Prob-BBDM method demonstrates superior preservation of anatomical structures and pathological features across all translation tasks, closely reaching the target images.}
    \label{fig:Results}
\end{figure*} 

\subsection{Quantitative comparisons}

\begin{table*}[t]
\centering
\caption{Quantitative comparison for the sequence-to-sequence translation task
on the BraTS 2021 dataset.
SSIM (\%) and PSNR (dB) are reported as mean $\pm$ std on the test set.}
\label{tbl:BraTSComparison}

\resizebox{\textwidth}{!}{%
\begin{tabular}{lcccccc}
\toprule
\multirow{2}{*}{\textbf{Model}} &
\multicolumn{2}{c}{\textbf{T1 $\rightarrow$ T2}} &
\multicolumn{2}{c}{\textbf{T2 $\rightarrow$ FLAIR}} &
\multicolumn{2}{c}{\textbf{FLAIR $\rightarrow$ T1}} \\
& \textbf{SSIM $\uparrow$} & \textbf{PSNR $\uparrow$}
& \textbf{SSIM $\uparrow$} & \textbf{PSNR $\uparrow$}
& \textbf{SSIM $\uparrow$} & \textbf{PSNR $\uparrow$} \\
\midrule
BBDM
& $61.22 \pm 3.57$ & $18.26 \pm 1.31$
& $64.47 \pm 3.57$ & $21.54 \pm 1.81$
& $69.04 \pm 3.74$ & $20.61 \pm 1.16$ \\

MaskGAN
& $74.29 \pm 2.18$ & $20.43 \pm 0.87$
& $69.05 \pm 5.96$ & $21.28 \pm 2.68$
& $78.89 \pm 2.98$ & $19.68 \pm 2.21$ \\

Pix2Pix
& $79.01 \pm 3.68$ & $22.65 \pm 1.70$
& $70.90 \pm 4.72$ & $21.46 \pm 2.94$
& $74.71 \pm 3.35$ & $21.38 \pm 2.19$ \\

ResViT
& $83.93 \pm 3.78$ & $24.18 \pm 1.76$
& $75.15 \pm 2.95$ & $22.64 \pm 2.43$
& $78.32 \pm 4.31$ & $22.15 \pm 2.46$ \\

SelfRDB
& $87.49 \pm 3.93$ & $24.76 \pm 2.30$
& $80.89 \pm 3.51$ & $23.51 \pm 2.48$
& $82.13 \pm 3.57$ & $20.76 \pm 2.38$ \\

\textbf{Prob-BBDM (Ours)}
& $\mathbf{88.46} \pm \mathbf{4.21}$ & $\mathbf{26.09} \pm \mathbf{1.94}$
& $\mathbf{83.89} \pm \mathbf{3.02}$ & $\mathbf{26.07} \pm \mathbf{1.76}$
& $\mathbf{85.11} \pm \mathbf{3.96}$ & $\mathbf{25.20} \pm \mathbf{1.98}$ \\
\bottomrule
\end{tabular}%
}
\end{table*}

\begin{figure*}
    \includegraphics[width=\linewidth]{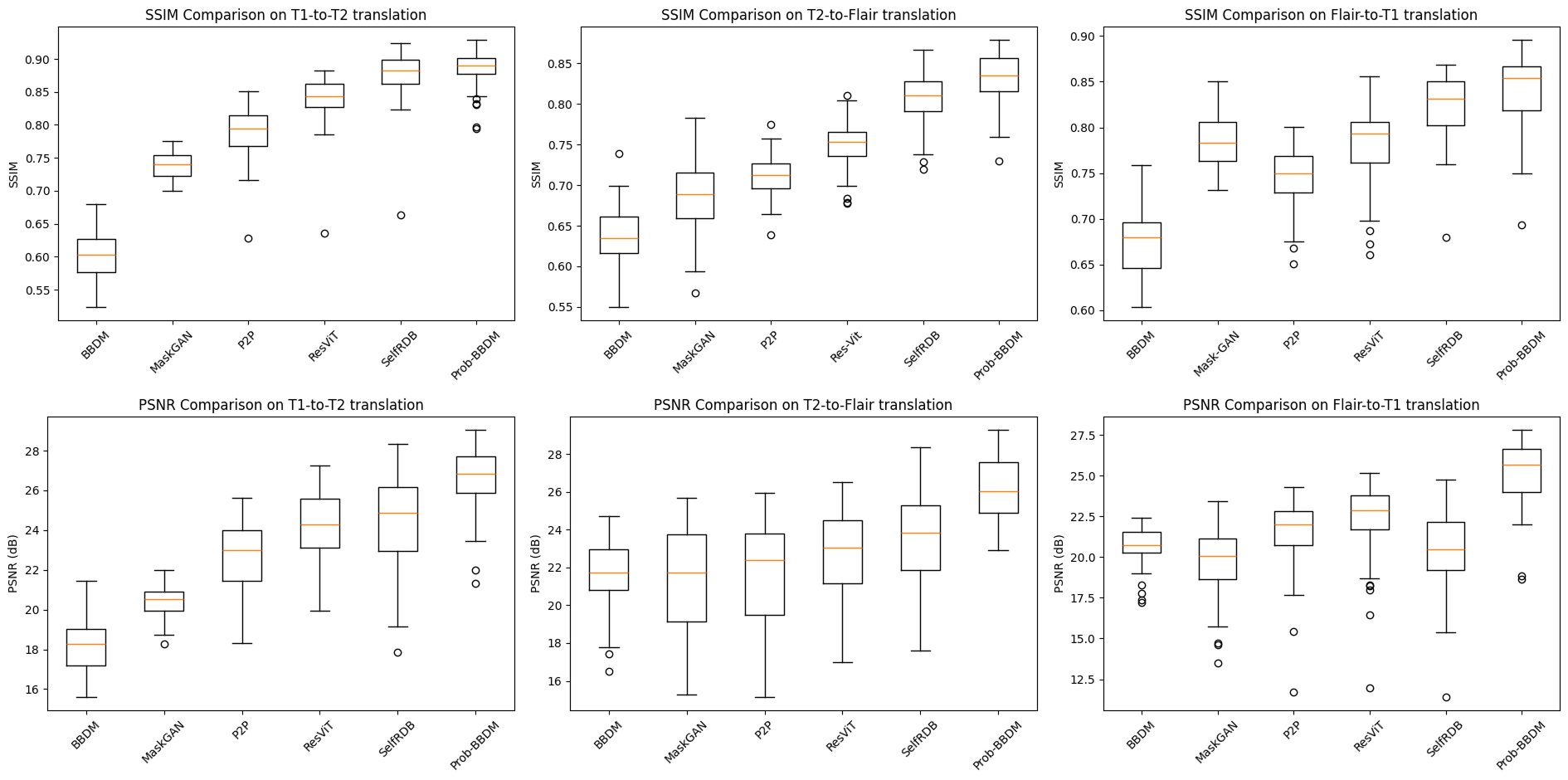}
    \centering
    \caption{Axial cross-sections slice-wise SSIM (top) and PSNR (bottom) distributions for T1-to-T2, T2-to-FLAIR, and FLAIR-to-T1 translation tasks. Prob-BBDM demonstrates consistently superior reconstruction fidelity, achieving the highest median values and reduced dispersion compared to baselines models. Statistical evaluation confirmed significant differences among models, with Holm-corrected Wilcoxon signed-rank tests indicating that Prob-BBDM significantly outperforms all competing approaches across all tasks.}
    \label{fig:Results Boxplots}
\end{figure*} 

Based on the numerical results presented in \autoref{tbl:BraTSComparison}, the proposed Prob-BBDM model consistently outperforms the competing methods across all translation tasks (T1-to-T2, T2-to-FLAIR, and FLAIR-to-T1), achieving the highest SSIM and PSNR values. While the absolute numerical gains may appear moderate, such improvements are meaningful in the context of MRI sequence translation, where SSIM is strongly correlated with the preservation of anatomical structures and tissue boundaries. 
In particular, average SSIM values above $85\%$, with peak average performance reaching $88.46\%$ for T1-to-T2 synthesis, indicate improved reconstruction of both normal anatomy and pathological regions, which is critical for maintaining clinical interpretability. Moreover, the consistent improvements observed across all translation tasks suggest that the proposed approach provides stable and robust performance rather than task-specific gains.
In terms of PSNR, our model achieves an average improvement of approximately $2.32\pm1.17 $dB over the best-performing baseline models in the three tasks. This consistent gain highlights the effectiveness of incorporating probabilistic modeling with a targeted conditioning mechanism, enabling more accurate reconstructions with reduced noise and artifact levels.
As illustrated in \autoref{fig:Results Boxplots}, Prob-BBDM consistently achieves the highest median SSIM and PSNR values across all translation tasks. The reduced interquartile range further indicates improved stability and robustness of the proposed method. These findings are statistically supported by Friedman ($p < 10^{-39}$) tests and Holm-corrected Wilcoxon signed-rank tests ($p < 0.05$), confirming significant improvements over all competing approaches. To further assess inter-slice coherence, we computed volumetric (3D) SSIM and PSNR across T2 reconstructed volumes from T1 sequence. The proposed method achieved a 3D SSIM of $85.1 \pm 4.1\%$ and a 3D PSNR of $24.43 \pm 2.23$ dB. In addition, we evaluated inter-slice consistency by measuring the similarity between consecutive axial slices. Compared to ground-truth volumes, the synthesized volumes exhibited a mean SSIM gap of $-0.036$ and a PSNR gap of $-1.48$ dB along the z-axis. This indicates a reduction in inter-slice coherence, while maintaining strong overall structural fidelity.

\begin{table}[h]
\centering
\caption{Computational comparison of state-of-the-art image translation models
in terms of number of trainable parameters and average inference time (ms).}
\label{tbl:ComputationalComparison}

\begin{tabular}{lll}
\toprule
\textbf{Model} & \textbf{\# Params.} & \textbf{Time (ms)} \\
\midrule
BBDM & 237.08M & 175.3 \\
MaskGAN & 11.78M & 45.35 \\
Pix2Pix & 54.41M & 15.01 \\
ResViT & 218.09M & 9.79 \\
SelfRDB & 44.1M & 1566.3 \\
\textbf{Prob-BBDM (Ours)} & 239.88M & 188.5 \\
\bottomrule
\end{tabular}

\end{table}

\subsection{Qualitative Results}
The results of our experiments on sequence-to-sequence translation are shown in \autoref{fig:Results} The Pix2Pix  produces effective translations by reconstructing a coherent anatomical structure in the cortical sulcus and ventricles. However, the results often appear pixelated, and the model struggles to reliably reconstruct the critical pathological areas.
Although the dual mechanism works well for reconstructing a sequence by cycle consistency \citep{zhu_unpaired_2020}. The MaskGAN model, originally designed for unpaired data translation, seems to focus too much on the pathological zone, given the generation of the mask. 
ResVit produces good results and could provide good alternatives, but shows lower quality in contrast synthesis. SelfRDB performs well for T1-to-T2 translation but struggles to reliably synthesize consistent contrast across modalities, especially in the vicinity of pathological regions, and exhibits a significantly higher inference time. 
Our Prob-BBDM model outperforms these approaches, producing visually high-quality samples with well-preserved anatomical structures. Specifically, the T1-to-T2 translation improves necrotic area delineation and enhancement. The T2-to-FLAIR translation effectively suppresses necrotic fluid components and artifacts while preserving the cortical ribbon. Meanwhile, the FLAIR-to-T1 translation achieves the best compromise between edema delineation and acquisition quality, particularly in the case of insulinoma tumors. As summarized in Table \autoref{tbl:ComputationalComparison}, Prob-BBDM has a computational cost comparable to BBDM and higher than lighter architectures such as Pix2Pix and MaskGAN, both in terms of parameter count and inference time. ResViT achieves a notably low inference time despite a parameter count comparable to Prob-BBDM; however, this computational efficiency comes at the expense of reduced contrast fidelity. In contrast, the increased complexity of Prob-BBDM remains moderate and enables substantially improved synthesis quality and robustness, particularly in pathological areas. Although SelfRDB has fewer parameters, its inference time is significantly higher, limiting its applicability in time-constrained clinical settings. Overall, these results highlight a favorable trade-off achieved by Prob-BBDM between synthesis quality and computational cost.

\subsection{Segmentation task}
\begin{figure*}
    \includegraphics[width=\linewidth]{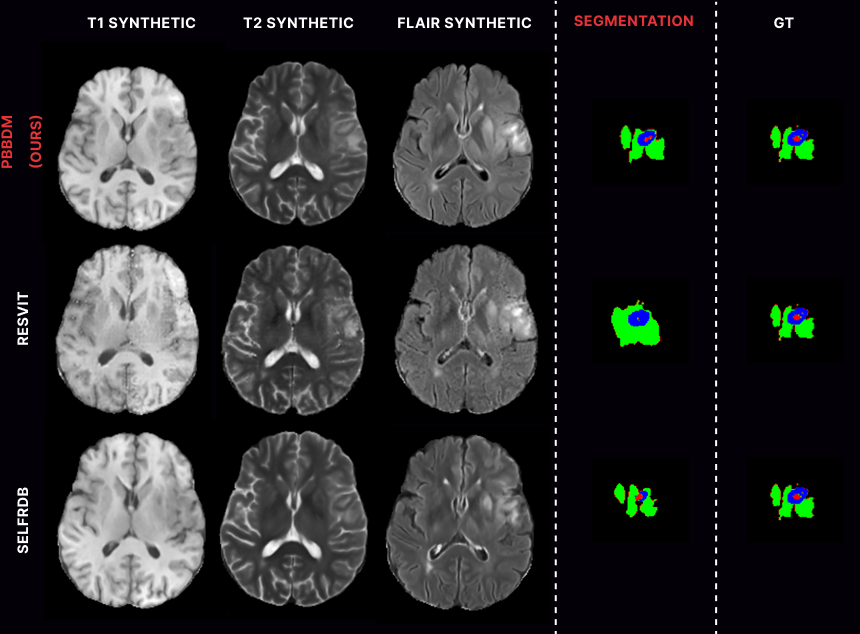}
    \centering
    \caption{Examples of multimodal MRI visualizations and associated segmentations. The first row shows synthetic slices generated by our model, along with corresponding segmentation masks predicted by nnUNet. The following rows show the results obtained using the ResViT (2nd row) and SelfRDB (3rd row) models as references. The first three columns display axial slices of synthetic T1, T2, and FLAIR MRIs, respectively. The last two columns show the predicted segmentation masks (SEGMENTATION) and the ground truth labels (GT) extracted from the original T1, T2, and FLAIR sequences of the BraTS 2021 dataset. The segmentation masks are color-coded: green for edema, blue for tumor, and red for necrosis.}
    \label{fig:ResultsSeg}
\end{figure*} 

\begin{table}[h]
\centering
\caption{Quantitative segmentation results on sequence-to-sequence translation
of T1, T2 and FLAIR synthesized from the test dataset.
DSC (\%) and HD95 (mm) are shown as mean $\pm$ std.}
\label{tbl:SegmentationMetrics}

\begin{tabular}{lll}
\toprule
\textbf{Model} & \textbf{DSC $\uparrow$} & \textbf{HD95 $\downarrow$} \\
\midrule
BBDM    & $85.68 \pm 22.07$ & $4.55 \pm 6.63$ \\
MaskGAN & $81.70 \pm 19.17$ & $8.91 \pm 9.40$ \\
Pix2Pix & $83.38 \pm 17.64$ & $8.84 \pm 11.79$ \\
ResViT  & $83.90 \pm 18.17$ & $8.53 \pm 11.81$ \\
SelfRDB & $88.23 \pm 16.22$ & $3.92 \pm 6.75$ \\
\textbf{Prob-BBDM (Ours)} & $\mathbf{88.71} \pm \mathbf{20.60}$ & $\mathbf{3.49} \pm \mathbf{4.72}$ \\
\midrule
GT slices & $90.38 \pm 19.05$ & $2.75 \pm 4.38$ \\
\bottomrule
\end{tabular}

\end{table}

Automatic segmentation is essential for the brain tumor diagnosis \citep{bakas} as it is challenging \citep{futrega_optimized_2021}. To further validate the effectiveness of our proposed Prob-BBDM translation model, we conducted a downstream task evaluation using the nnU-Net segmentation framework \citep{luu_extending_2021}. We input the synthesized MRI sequences (T1, T2, FLAIR) generated by each model into a pre-trained nnU-Net v2 model trained on BraTS 2021. The model uses a 5-level U-Net with 3×3×3 convolutions (3D), LeakyReLU activations, and Instance Normalization. It was trained with soft Dice + cross-entropy loss, SGD optimizer with momentum 0.99, polynomial learning rate decay, and extensive data augmentation including rotations, scaling, mirroring, gamma adjustments, and noise. 
This strategy provides an indirect yet practical evaluation of the anatomical fidelity and realism of the synthesized images, as segmentation performance is highly sensitive to structural consistency and contrast quality. \\
Qualitatively, our model’s synthetic modalities lead to segmentations that exhibit slightly better delineation of the necrotic and enhancing tumor regions (red and blue), with improved boundary sharpness and spatial alignment to the ground truth compared to ResViT. The edema regions (green) are similarly captured in both methods, although our model shows fewer false positives in surrounding tissues. To further ensure the reliability and clinical relevance of these findings, all resulting segmentations were reviewed and validated by a senior radiologist. All examination were blindly reviewed in DICOM Format on the same Workstation (Change Healthcare, Chicago, USA). 

As shown in \autoref{tbl:SegmentationMetrics}, our method achieves the highest DSC of $88.71 \pm 20.60\%$ and the lowest HD95 of $3.49 \pm 4.72mm$, indicating more accurate and spatially consistent tumor delineations compared to other synthetic approaches. To assess statistical significance, we conducted paired Wilcoxon signed-rank tests comparing Prob-BBDM with competing methods. For HD95, Prob-BBDM significantly outperformed BBDM, MaskGAN, Pix2Pix, and ResViT ($p < 0.001$), while no statistically significant difference was observed compared to SelfRDB ($p = 0.186$) indicating competitive boundary accuracy with the strongest diffusion based methods. For DSC, Prob-BBDM achieved statistically significant improvements over all baselines, including SelfRDB ($p < 0.001$). This suggests a consistent improvement in overlap accuracy, highlighting the robustness of the proposed framework.

\subsection{Validation}

\begin{table}[h]
\centering
\caption{Quantitative results for sequence-to-sequence translation on the Gliobiopsy dataset.
SSIM (\%) and PSNR (dB) are shown as mean $\pm$ std.}
\label{tbl:GliobiopsyMetrics}
\begin{tabular}{llll}
\toprule
\textbf{Training Set} & \textbf{Test Set} & \textbf{SSIM $\uparrow$} & \textbf{PSNR $\uparrow$} \\
\midrule
1250 Patients & 59 Patients & $86.11 \pm 4.64$ & $23.6 \pm 1.81$ \\
\bottomrule
\end{tabular}
\end{table}

To evaluate the generalization capability of our proposed Prob-BBDM, we conducted an external validation study using an internal, non-public dataset collected at our affiliated institution. 
This dataset was entirely excluded from model training and fine-tuning, providing a fully independent test set to assess real-world performance. The test cohort included 59 patients (median age: 50 years; interquartile range: 36–61 years; 29 females) with suspected glioma, who underwent preoperative brain MRI at Poitiers University Hospital. Imaging was performed on a 3T Skyra system (Siemens, Erlangen, Germany), and included sagittal 3D FLAIR and axial 3D T1 MPRAGE sequences. In this study, we focused on the FLAIR-to-T1 translation task using our proposed Prob-BBDM. The synthetic T1-weighted (T1w) images were evaluated using the SSIM and PSNR, qualitative edema delineation and expert visual assessment using a 5-point Likert scale. 
The Prob-BBDM model was trained exclusively on paired FLAIR–T1w scans from the BraTS 2021 dataset, comprising 1250 patients from multiple institutions and scanner vendors (Siemens, GE Healthcare, and Philips). Notably, this training set included a mix of 1.5T and 3T images, whereas the test set consisted solely of 3T MRI.
Despite this domain gap, our model achieved strong performance on the external test set, with a mean SSIM of $86.11\pm4.64\%$ and a PSNR of $23.6\pm1.81 $dB (see \autoref{tbl:GliobiopsyMetrics}). 

\begin{figure}
    \centering
    \includegraphics[width=\columnwidth]{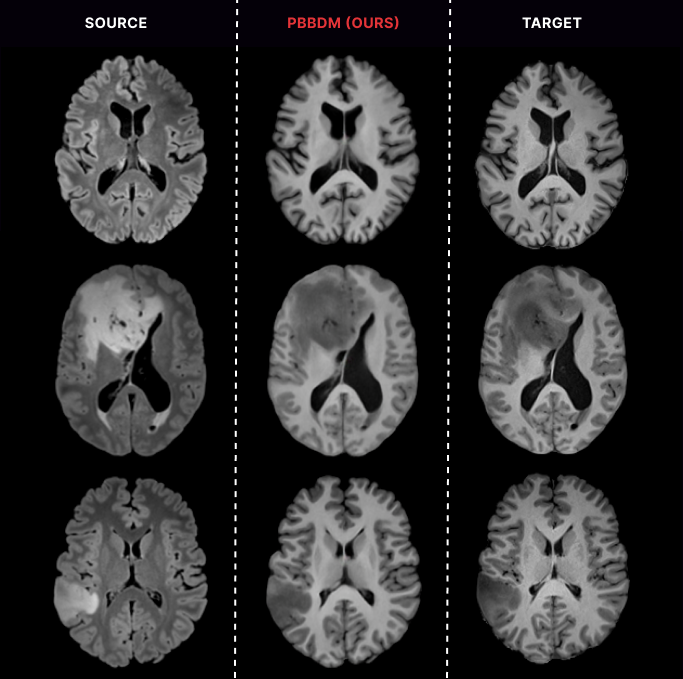}
    \caption{Qualitative results of synthesized 3T MRI sequences using Prob-BBDM. (1st row) Healthy slice, (2nd row) Glioblastoma, and (3rd row) Astrocytoma. From left to right: source image, outputs from our Prob-BBDM, and the ground-truth target image.}
    \label{fig:Results_Glio}
\end{figure}

These results indicate that the model retains high structural fidelity and generalizes well to unseen clinical data.
Qualitative assessment further supports these findings. As illustrated in \autoref{fig:Results_Glio}, the synthetic T1w images preserve key anatomical features such as white/gray matter contrast and edema boundaries, showing strong consistency with native T1w images. Moreover, expert visual assessment using a 5-point Likert scale confirmed that the synthetic images enabled accurate interpretation of contrast enhancement patterns on corresponding Gadolinium-enhanced scans. Together, these results demonstrate that Prob-BBDM can generate high-fidelity synthetic T1w images across domains, supporting its potential integration into clinical neuroimaging workflows.

\subsection{Ablation Study}
We conducted a series of ablation experiments to evaluate the contribution of key components in our model.
All experiments were performed on the same T1-to-T2 sequence translation task to ensure consistency and comparability.

\begin{table}[h]
\centering
\caption{Performance of our model variants ablated of $y$ source-image guidance,
and modified Prob-UNet with KL distance matching and weighting factor.
SSIM (\%) and PSNR (dB) are reported as mean $\pm$ std on the test set.}
\label{tbl:AblationStudy}

\begin{tabular}{lll}
\toprule
\textbf{Method} & \textbf{SSIM $\uparrow$} & \textbf{PSNR $\uparrow$} \\
\midrule
BBDM & $72.78 \pm 3.66$ & $21.88 \pm 1.43$ \\
$+\,D_{\mathrm{KL}}$ & $87.32 \pm 3.95$ & $25.94 \pm 1.85$ \\
$+\,\lambda\,D_{\mathrm{KL}}$ & $\mathbf{88.59} \pm \mathbf{3.73}$ & $\mathbf{26.47} \pm \mathbf{1.89}$ \\
\bottomrule
\end{tabular}

\end{table}

\begin{table}[h]
\centering
\caption{Performance of our model variants with different values of the
weighting factor $\lambda$.
SSIM (\%) and PSNR (dB) are reported as mean $\pm$ std on the test set.}
\label{tbl:LambdaValueStudy}

\begin{tabular}{lll}
\toprule
$\boldsymbol{\lambda}$ & \textbf{SSIM $\uparrow$} & \textbf{PSNR $\uparrow$} \\
\midrule
$10^{-1}$ & $87.32 \pm 3.80$ & $26.09 \pm 1.64$ \\
$10^{-2}$ & $87.64 \pm 3.79$ & $26.46 \pm 1.76$ \\
$10^{-3}$ & $\mathbf{88.59} \pm \mathbf{3.73}$ & $\mathbf{26.47} \pm \mathbf{1.89}$ \\
$10^{-4}$ & $87.92 \pm 3.78$ & $26.09 \pm 1.64$ \\
\bottomrule
\end{tabular}

\end{table}

\subsubsection{Influence of the Probabilistic setting}
To investigate the effectiveness of the probabilistic loss and the impact of the weighting factor, we conducted an ablation study using the BBDM model as a baseline. We progressively incorporated the probabilistic loss term (denoted as $D_{KL}$) and then introduced the weighting factor $\lambda$ to assess their individual contributions. As shown in \autoref{tbl:AblationStudy}, adding the $D_{KL}$ term led to a substantial improvement in both SSIM (from $72.78\%$ to $87.51\%$) and PSNR (from $21.88 $dB to $26.24 $dB), indicating that the probabilistic formulation significantly enhances image quality. Introducing the weighting factor $\lambda$ further improved the results to an SSIM of $88.59\%$ and a PSNR of $26.40 $dB, suggesting that balancing the influence of the KL divergence term leads to more stable and effective training.
Overall, pairwise Wilcoxon signed-rank tests with Holm correction indicate that Prob-BBDM outperforms the ablated variants in terms of translation performance ($p < 0.05$).

To further examine the influence of the weighting factor $\lambda$, we performed an additional ablation where we varied its value over four orders of magnitude. As shown in \autoref{tbl:LambdaValueStudy}, performance was sensitive to the choice of  $\lambda$. A value of  $\lambda=10^{-3}$ yielded the highest SSIM ($88.59\%$) and a near-optimal PSNR ($26.47 $dB), indicating a favorable balance between the data reconstruction loss and the KL divergence. Larger values (\textit{e.g.}, $10^{-1}$) and smaller values (\textit{e.g.}, $10^{-4}$) resulted in slightly lower performance, suggesting that both under and over regularization can negatively impact synthesis quality.

\begin{table}[h]
\centering
\caption{Performance of our model as a function of the number of sampling steps
used during the reverse process.
SSIM (\%) and PSNR (dB) are reported as mean $\pm$ std on the test set.
Average inference time (IT) per patient is reported in milliseconds.}
\label{tbl:SamplingStepStudy}
\begin{tabular}{llll}
\toprule
\textbf{Sampling steps} & \textbf{SSIM $\uparrow$} & \textbf{PSNR $\uparrow$} & \textbf{IT (ms)} \\
\midrule
2   & $87.32 \pm 4.48$ & $25.26 \pm 2.14$ & 92.9 \\
4   & $\mathbf{88.18} \pm \mathbf{4.21}$ & $\mathbf{26.12} \pm \mathbf{2.02}$ & 188.5 \\
20  & $81.22 \pm 5.53$ & $22.49 \pm 2.03$ & 945.3 \\
100 & $76.41 \pm 5.83$ & $21.13 \pm 1.97$ & 4803.1 \\
200 & $75.77\pm 5.86$ & $21.01 \pm 1.90$ & 7328.4 \\
\bottomrule
\end{tabular}
\end{table}

\subsubsection{Influence of Sampling Steps}
\begin{figure}
    \centering
    \includegraphics[width=\columnwidth]{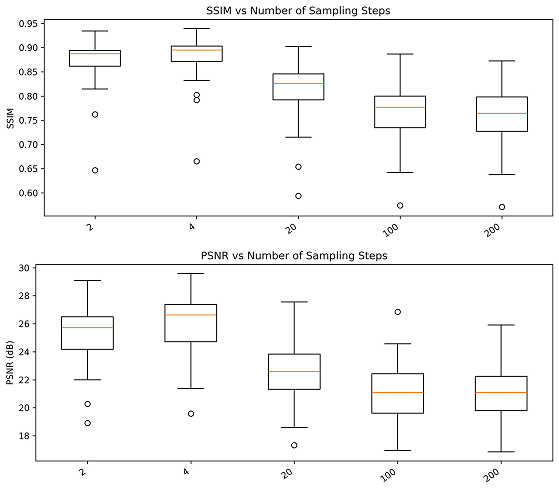}
    \caption{Slice-wise SSIM (top) and PSNR (bottom) distributions across test subjects for varying numbers of diffusion sampling steps. The 4 steps configuration achieves the highest median performance and improved stability. Statistical analysis confirmed significant differences among configurations with Friedman and Holm-corrected Wilcoxon signed-rank tests.}

    \label{fig:Boxplots Ablation Study}
\end{figure}
To evaluate the impact of the number of diffusion sampling steps, we conducted a comprehensive ablation study summarized in \autoref{tbl:SamplingStepStudy}. All experiments were performed using the finalized training configuration and identical evaluation protocol to ensure consistency across settings. The best results were obtained using $4$ sampling steps, reaching an average SSIM of $88.18 \pm 4.21$ and a PSNR of $26.123 \pm 2.02$ dB.
Increasing the number of steps resulted in a marked degradation. In particular, performance progressively declined for 20, 100, and 200 steps, with SSIM dropping to $75.77$ and PSNR to $21.01$ dB at 200 steps. This degradation suggests that excessive denoising within the Brownian bridge framework may introduce smoothing artifacts and cause the reverse trajectory to drift away from the optimal translation path. To statistically validate these observations, we performed a Friedman test across all step configurations. The differences were highly significant for both SSIM ($p < 10^{-37}$) and PSNR ($p < 10^{-38}$). Post-hoc pairwise Wilcoxon signed-rank tests with Holm correction confirmed that the $4$ steps configuration significantly outperformed all other settings (all corrected $p < 0.001$). 
The corresponding box plots (\autoref{fig:Boxplots Ablation Study}) illustrate not only higher median performance for four sampling steps but also a tighter distribution compared to higher-step configurations, indicating stable behavior across subjects. In addition to improved image quality, reducing the number of sampling steps substantially lowers computational cost and inference time. This favorable trade-off between efficiency and reconstruction fidelity makes Prob-BBDM particularly suitable for clinical workflows where rapid processing is required.

\section{Discussion}
Prob-BBDM is a diffusion-based approach for medical image translation, designed to progressively transform an image from a source modality to a target modality. By leveraging a Brownian bridge formulation, the diffusion process is explicitly constrained between the known source and target endpoints, enforcing a smooth and probabilistically translation path.

In contrast to standard diffusion models, which are typically optimized for sample diversity and require exploring a large image space starting from random noise, the Brownian bridge formulation substantially restricts the diffusion trajectory to an informative subspace. This structural constraint is the primary factor enabling Prob-BBDM to achieve high-quality results with a very small number of sampling steps. As a result, accurate translations can be obtained with only 2 to 4 diffusion steps, whereas conventional diffusion models often require hundreds or thousands of iterations.

The integration of probabilistic conditioning through variational encoders further stabilizes the diffusion process in this low-step configuration. While the Brownian bridge provides the acceleration mechanism, the learned prior and posterior distributions reduce uncertainty during denoising and prevent unstable trajectories that may arise when the number of diffusion steps is limited. Together, these components shift the focus from sample diversity toward higher structural similarity and clinical relevance, enabling synthesized images to match the target modality while preserving anatomical details.
The ablation study (\autoref{tbl:SamplingStepStudy}) confirms that Prob-BBDM maintains strong performance with as few as 2 to 4 diffusion steps, achieving a favorable trade-off between efficiency and image quality. This reduction in sampling steps leads to significantly lower computational cost and faster inference, which is particularly important for practical and clinical applications.

\par Although recent diffusion-based approaches have demonstrated promising results for 3D medical image synthesis, most diffusion models for image-to-image translation are still formulated and evaluated in 2D settings. This choice is primarily driven by the substantial computational and memory demands of 3D diffusion models, which often require significantly reduced resolution, limited batch sizes, or extensive hardware resources to remain tractable.
In the context of MRI sequence translation, 2D slice-based processing remains a widely adopted and effective strategy, allowing high-resolution modeling while preserving fine anatomical details. Although a slight reduction in inter-slice coherence was observed compared to ground-truth volumes, the magnitude of the z-axis gap remained moderate. These findings suggest that slice-wise processing preserves volumetric structure reasonably well, while leaving room for further improvements through explicit cross slice modeling. Moreover, the Brownian bridge formulation used in this work focuses on learning accurate source-to-target translations at the image level, where structural fidelity within each slice is critical. While explicit 3D volumetric consistency is not directly enforced in the current framework, the use of anatomically aligned slices and standardized preprocessing mitigates major inter-slice discontinuities. Nevertheless, purely slice-wise processing may still introduce subtle inconsistencies along the z-axis.

Several strategies could enhance volumetric coherence without incurring the full computational burden of 3D diffusion models. A practical realistic approach is a 2.5D formulation, where adjacent slices are incorporated as additional input channels to provide local contextual information while preserving 2D computational efficiency. Another promising direction involves modeling slice-level dependencies in latent space using autoregressive or attention-based mechanisms. In such a formulation, each slice can be represented as a latent token derived from the diffusion backbone, and inter-slice continuity can be reinforced through sequential modeling across slices.
Extending the proposed Prob-BBDM framework to fully 3D volumetric diffusion models represents an important direction for future work, but is beyond the scope of the present study due to computational constraints.

Our model maintains a strong balance between synthesis quality and inference time, some outputs still exhibit a smoothing effect, which can diminish fine-grained anatomical details. A more detailed analysis of model behavior reveals that the main failure cases of Prob-BBDM occur in anatomically and radiologically challenging situations. Flow void artifacts may alter local intensity distributions and distort anatomical boundaries, which can affect contrast translation and edge preservation in tumors adjacent to cerebrospinal fluid spaces. Importantly, these challenging cases correspond to well-known difficulties in routine radiological interpretation. Low contrast-to-noise ratio, partial volume effects, and flow-related artifacts frequently complicate human MRI reading. Therefore, the observed failure modes of Prob-BBDM reflect intrinsic limitations of MRI signal characteristics rather than purely algorithmic instability. Finally, the probabilistic alignment between prior and posterior latent distributions, while improving stability and anatomical coherence, may introduce bias toward the learned training distribution. In rare or atypical pathological presentations, this may lead the model to generate anatomically plausible but clinically incomplete representations of small lesions. Consequently, synthesized images should be interpreted as complementary information rather than direct substitutes for fully acquired sequence

While Prob-BBDM is computationally efficient at low diffusion steps, it remains constrained in its ability to synthesize high-resolution outputs. Tackling high-resolution generation possibly through multi-scale diffusion strategies, patch-based synthesis, or memory-efficient training schemes will be essential for translating this approach to practical. High-resolution clinical imaging tasks such as translation 1.5T to 3T MRI can also be targeted, as shown by the results obtained on the Gliobiopsy dataset (\autoref{tbl:GliobiopsyMetrics}, \autoref{fig:Results_Glio}).
Together, these improvements could further broaden the utility of Prob-BBDM and push the boundaries of bridge diffusion based medical image translation.

\section{Conclusion}
We have presented a probabilistic diffusion-based model for MRI sequence translation that has the potential to accelerate imaging protocols, particularly benefiting patient populations such as agitated, confused, or pediatric individuals, without requiring changes to existing acquisition protocols. As a post-processing solution, our model preserves clinically relevant anatomical and pathological information, thereby supporting both patient management and ongoing research in synthetic medical data and digital twin technologies.
Our experiments demonstrate that the proposed model delivers rapid and high-quality image synthesis compared to existing diffusion approaches. While there remains room for improvement, particularly in accurately synthesizing boundary regions, our results establish a strong foundation for future work. This includes extending the model to incorporate contrast-enhanced sequences (\textit{e.g.}, T1ce) relevant to neuro-oncology, investigating applications in degenerative diseases, and integrating advanced MRI modalities such as perfusion and diffusion imaging.
Moreover, implementing our approach in a latent space framework such as a latent diffusion model \citep{rombach_high-resolution_2022} could significantly reduce computational requirements and enable scaling to larger datasets, ultimately enhancing the fidelity and clinical applicability of synthetic MRI translation.

\bibliographystyle{plainnat}
\bibliography{References}

\begin{thebibliography}{43}
\providecommand{\natexlab}[1]{#1}
\providecommand{\url}[1]{\texttt{#1}}
\expandafter\ifx\csname urlstyle\endcsname\relax
  \providecommand{\doi}[1]{doi: #1}\else
  \providecommand{\doi}{doi: \begingroup \urlstyle{rm}\Url}\fi

\bibitem[AIRS(2025)]{airs_swiftmr_2025}
AIRS.
\newblock Swiftmr™ product information | ai-powered mri enhancement solution
  – airs medical inc.
\newblock \url{https://airsmed.com/swiftmr/}, 2025.

\bibitem[Armanious et~al.(2020)Armanious, Jiang, Fischer, Küstner, Hepp,
  Nikolaou, Gatidis, and Yang]{armanious_medgan_2019}
Karim Armanious, Chenming Jiang, Marc Fischer, Thomas Küstner, Tobias Hepp,
  Konstantin Nikolaou, Sergios Gatidis, and Bin Yang.
\newblock {MedGAN}: Medical image translation using {GANs}.
\newblock \emph{Computerized Medical Imaging and Graphics}, 79:\penalty0
  101684, 2020.
\newblock ISSN 0895-6111.
\newblock \doi{https://doi.org/10.1016/j.compmedimag.2019.101684}.
\newblock URL
  \url{https://www.sciencedirect.com/science/ARTICLE/pii/S0895611119300990}.

\bibitem[Arslan et~al.(2025)Arslan, Kabas, Dalmaz, Ozbey, and
  Çukur]{arslan_self-consistent_2025}
Fuat Arslan, Bilal Kabas, Onat Dalmaz, Muzaffer Ozbey, and Tolga Çukur.
\newblock Self-consistent recursive diffusion bridge for medical image
  translation.
\newblock \emph{Medical Image Analysis}, 106:\penalty0 103747, 2025.
\newblock ISSN 1361-8415.
\newblock \doi{10.1016/j.media.2025.103747}.
\newblock URL
  \url{https://www.sciencedirect.com/science/article/pii/S1361841525002944}.

\bibitem[Atli et~al.(2024)Atli, Kabas, Arslan, Yurt, Dalmaz, and
  Çukur]{atli_i2i-mamba_2025}
Omer~F. Atli, Bilal Kabas, Fuat Arslan, Mahmut Yurt, Onat Dalmaz, and Tolga
  Çukur.
\newblock I2i-mamba: Multi-modal medical image synthesis via selective state
  space modeling.
\newblock \emph{arXiv:2405.14022}, 2024.

\bibitem[Cao et~al.(2024)Cao, Tan, Gao, Xu, Chen, Heng, and
  Li]{cao_survey_2023}
Hanqun Cao, Cheng Tan, Zhangyang Gao, Yilun Xu, Guangyong Chen, Pheng-Ann Heng,
  and Stan~Z. Li.
\newblock A survey on generative diffusion models.
\newblock \emph{IEEE Transactions on Knowledge \& Data Engineering},
  36\penalty0 (07):\penalty0 2814--2830, 2024.
\newblock \doi{10.1109/TKDE.2024.3361474}.

\bibitem[Choi et~al.(2021)Choi, Kim, Jeong, Gwon, and Yoon]{choi_ilvr_2021}
Jooyoung Choi, Sungwon Kim, Yonghyun Jeong, Youngjune Gwon, and Sungroh Yoon.
\newblock {ILVR}: Conditioning method for denoising diffusion probabilistic
  models.
\newblock In \emph{2021 {IEEE}/{CVF} International Conference on Computer
  Vision ({ICCV})}, pages 14347--14356. {IEEE}, 2021.
\newblock ISBN 978-1-66542-812-5.

\bibitem[Dalmaz et~al.(2022)Dalmaz, Yurt, and Çukur]{dalmaz_resvit_2022}
Onat Dalmaz, Mahmut Yurt, and Tolga Çukur.
\newblock {ResViT}: Residual vision transformers for multi-modal medical image
  synthesis.
\newblock \emph{{IEEE} Transactions on Medical Imaging}, 41\penalty0
  (10):\penalty0 2598--2614, 2022.
\newblock ISSN 0278-0062, 1558-254X.
\newblock \doi{10.1109/TMI.2022.3167808}.
\newblock URL \url{http://arxiv.org/abs/2106.16031}.

\bibitem[Dayarathna et~al.(2024)Dayarathna, Islam, Uribe, Yang, Hayat, and
  Chen]{dayarathna_deep_2024}
Sanuwani Dayarathna, Kh~Tohidul Islam, Sergio Uribe, Guang Yang, Munawar Hayat,
  and Zhaolin Chen.
\newblock Deep learning based synthesis of {MRI}, {CT} and {PET}: Review and
  analysis.
\newblock \emph{Medical Image Analysis}, 92:\penalty0 103046, 2024.
\newblock ISSN 13618415.
\newblock \doi{10.1016/j.media.2023.103046}.

\bibitem[Dhariwal and Nichol(2021)]{dhariwal_diffusion_2021}
Prafulla Dhariwal and Alexander~Quinn Nichol.
\newblock Diffusion models beat {GAN}s on image synthesis.
\newblock In A.~Beygelzimer, Y.~Dauphin, P.~Liang, and J.~Wortman Vaughan,
  editors, \emph{Advances in Neural Information Processing Systems}, 2021.
\newblock URL \url{https://openreview.net/forum?id=AAWuCvzaVt}.

\bibitem[Dosovitskiy et~al.(2020)Dosovitskiy, Beyer, Kolesnikov, Weissenborn,
  Zhai, Unterthiner, Dehghani, Minderer, Heigold, Gelly,
  et~al.]{dosovitskiy_image_2021}
Alexey Dosovitskiy, Lucas Beyer, Alexander Kolesnikov, Dirk Weissenborn,
  Xiaohua Zhai, Thomas Unterthiner, Mostafa Dehghani, Matthias Minderer, Georg
  Heigold, Sylvain Gelly, et~al.
\newblock An image is worth 16x16 words: Transformers for image recognition at
  scale.
\newblock \emph{arXiv preprint arXiv:2010.11929}, 2020.

\bibitem[Futrega et~al.(2022)Futrega, Milesi, Marcinkiewicz, and
  Ribalta]{futrega_optimized_2021}
Michał Futrega, Alexandre Milesi, Michał Marcinkiewicz, and Pablo Ribalta.
\newblock Optimized u-net for brain tumor segmentation, 2022.

\bibitem[Goodfellow et~al.(2014)Goodfellow, Pouget-Abadie, Mirza, Xu,
  Warde-Farley, Ozair, Courville, and Bengio]{goodfellow_generative_2014}
Ian Goodfellow, Jean Pouget-Abadie, Mehdi Mirza, Bing Xu, David Warde-Farley,
  Sherjil Ozair, Aaron Courville, and Yoshua Bengio.
\newblock Generative adversarial nets.
\newblock In Z.~Ghahramani, M.~Welling, C.~Cortes, N.~Lawrence, and K.~Q.
  Weinberger, editors, \emph{Advances in Neural Information Processing
  Systems}, volume~27. Curran Associates, Inc., 2014.
\newblock URL
  \url{https://proceedings.neurips.cc/paper_files/paper/2014/file/5ca3e9b122f61f8f06494c97b1afccf3-Paper.pdf}.

\bibitem[Goswami et~al.(2025)Goswami, Gupta, and Paul]{goswami_learndiff_2025}
Sagnik Goswami, Akriti Gupta, and Angshuman Paul.
\newblock {LearnDiff}: {MRI} image super-resolution using a diffusion model
  with learnable noise.
\newblock \emph{Computerized Medical Imaging and Graphics}, 125:\penalty0
  102641, 2025.
\newblock ISSN 0895-6111.
\newblock \doi{10.1016/j.compmedimag.2025.102641}.
\newblock URL
  \url{https://www.sciencedirect.com/science/ARTICLE/pii/S0895611125001508}.

\bibitem[Guo et~al.(2025)Guo, Han, Lyu, Zhou, and Shen]{guo_towards_2025}
Dongqian Guo, Wencheng Han, Pang Lyu, Yuxi Zhou, and Jianbing Shen.
\newblock Towards better cephalometric landmark detection with diffusion data
  generation.
\newblock \emph{IEEE Transactions on Medical Imaging}, 44\penalty0
  (7):\penalty0 2784--2794, 2025.

\bibitem[Ho et~al.(2020)Ho, Jain, and Abbeel]{ho_denoising_2020}
Jonathan Ho, Ajay Jain, and Pieter Abbeel.
\newblock Denoising diffusion probabilistic models.
\newblock In H.~Larochelle, M.~Ranzato, R.~Hadsell, M.F. Balcan, and H.~Lin,
  editors, \emph{Advances in Neural Information Processing Systems}, volume~33,
  pages 6840--6851. Curran Associates, Inc., 2020.

\bibitem[Isola et~al.(2017)Isola, Zhu, Zhou, and Efros]{pix2pix2017}
Phillip Isola, Jun-Yan Zhu, Tinghui Zhou, and Alexei~A Efros.
\newblock Image-to-image translation with conditional adversarial networks.
\newblock \emph{CVPR}, 2017.

\bibitem[Kazerouni et~al.(2023)Kazerouni, Aghdam, Heidari, Azad, Fayyaz,
  Hacihaliloglu, and Merhof]{kazerouni_diffusion_2023}
Amirhossein Kazerouni, Ehsan~Khodapanah Aghdam, Moein Heidari, Reza Azad,
  Mohsen Fayyaz, Ilker Hacihaliloglu, and Dorit Merhof.
\newblock Diffusion models in medical imaging: A comprehensive survey.
\newblock \emph{Medical Image Analysis}, 88:\penalty0 102846, 2023.
\newblock ISSN 1361-8415.
\newblock \doi{https://doi.org/10.1016/j.media.2023.102846}.

\bibitem[Kebaili et~al.(2025)Kebaili, Lapuyade-Lahorgue, Vera, and
  Ruan]{kebaili_multi-modal_2025}
Aghiles Kebaili, Jérôme Lapuyade-Lahorgue, Pierre Vera, and Su~Ruan.
\newblock Multi-modal {MRI} synthesis with conditional latent diffusion models
  for data augmentation in tumor segmentation.
\newblock \emph{Computerized Medical Imaging and Graphics}, 123:\penalty0
  102532, 2025.
\newblock ISSN 0895-6111.
\newblock \doi{10.1016/j.compmedimag.2025.102532}.
\newblock URL
  \url{https://www.sciencedirect.com/science/ARTICLE/pii/S0895611125000412}.

\bibitem[Khader et~al.(2023)Khader, Müller-Franzes, Tayebi~Arasteh, Han,
  Haarburger, Schulze-Hagen, Schad, Engelhardt, Baeßler, Foersch, Stegmaier,
  Kuhl, Nebelung, Kather, and Truhn]{khader_medical_2023}
Firas Khader, Gustav Müller-Franzes, Soroosh Tayebi~Arasteh, Tianyu Han,
  Christoph Haarburger, Maximilian Schulze-Hagen, Philipp Schad, Sandy
  Engelhardt, Bettina Baeßler, Sebastian Foersch, Johannes Stegmaier,
  Christiane Kuhl, Sven Nebelung, Jakob~Nikolas Kather, and Daniel Truhn.
\newblock Denoising diffusion probabilistic models for 3d medical image
  generation.
\newblock \emph{Scientific Reports}, 13\penalty0 (1):\penalty0 7303, 2023.
\newblock ISSN 2045-2322.
\newblock \doi{10.1038/s41598-023-34341-2}.
\newblock URL \url{https://doi.org/10.1038/s41598-023-34341-2}.

\bibitem[Kim and Park(2024)]{kim_adaptive_2024}
Jonghun Kim and Hyunjin Park.
\newblock Adaptive latent diffusion model for 3d medical image to image
  translation: Multi-modal magnetic resonance imaging study.
\newblock In \emph{Proceedings of the IEEE/CVF Winter Conference on
  Applications of Computer Vision (WACV)}, pages 7604--7613, January 2024.

\bibitem[Kohl et~al.(2019)Kohl, Romera-Paredes, Meyer, De~Fauw, Ledsam,
  Maier-Hein, Eslami, Rezende, and Ronneberger]{kohl_probabilistic_2019}
Simon~AA Kohl, Bernardino Romera-Paredes, Clemens Meyer, Jeffrey De~Fauw,
  Joseph~R Ledsam, Klaus~H Maier-Hein, SM~Eslami, Danilo~Jimenez Rezende, and
  Olaf Ronneberger.
\newblock A probabilistic u-net for segmentation of ambiguous images.
\newblock \emph{arXiv preprint arXiv:1806.05034}, 2019.

\bibitem[Kong et~al.(2021)Kong, Lian, Huang, Li, Hu, and
  Zhou]{kong_breaking_nodate}
Lingke Kong, Chenyu Lian, Detian Huang, Zhenjiang Li, Yanle Hu, and Qichao
  Zhou.
\newblock Breaking the dilemma of medical image-to-image translation.
\newblock In \emph{Proceedings of the 35th International Conference on Neural
  Information Processing Systems}, NIPS '21. Curran Associates Inc., 2021.
\newblock ISBN 9781713845393.

\bibitem[Li et~al.(2023)Li, Xue, Liu, and Lai]{li_bbdm_2023}
Bo~Li, Kaitao Xue, Bin Liu, and Yu-Kun Lai.
\newblock Bbdm: Image-to-image translation with brownian bridge diffusion
  models.
\newblock In \emph{Proceedings of the IEEE/CVF Conference on Computer Vision
  and Pattern Recognition}, pages 1952--1961, 2023.

\bibitem[Liu et~al.(2023)Liu, Pasumarthi, Duffy, Gong, Datta, and
  Zaharchuk]{liu_one_2023}
Jiang Liu, Srivathsa Pasumarthi, Ben Duffy, Enhao Gong, Keshav Datta, and Greg
  Zaharchuk.
\newblock One model to synthesize them all: Multi-contrast multi-scale
  transformer for missing data imputation.
\newblock \emph{{IEEE} Transactions on Medical Imaging}, 42\penalty0
  (9):\penalty0 2577--2591, 2023.
\newblock ISSN 0278-0062, 1558-254X.
\newblock \doi{10.1109/TMI.2023.3261707}.

\bibitem[Luu and Park(2022)]{luu_extending_2021}
Huan~Minh Luu and Sung-Hong Park.
\newblock Extending nn-{UNet} for brain tumor segmentation.
\newblock In Alessandro Crimi and Spyridon Bakas, editors, \emph{Brainlesion:
  Glioma, Multiple Sclerosis, Stroke and Traumatic Brain Injuries}, pages
  173--186. Springer International Publishing, 2022.
\newblock ISBN 978-3-031-09002-8.

\bibitem[Mohan et~al.(2021)Mohan, Bilello, Calabrese, Colak, and
  Farahani]{mohan_rsna-asnr-miccai_2021}
S.~Mohan, M.~Bilello, E.~Calabrese, E.~Colak, and K.~Farahani.
\newblock The {RSNA}-{ASNR}-{MICCAI} {BraTS} 2021 benchmark on brain tumor
  segmentation and radiogenomic classification, 2021.
\newblock URL \url{http://arxiv.org/abs/2107.02314}.

\bibitem[O'Shea and Nash(2015)]{oshea_introduction_2015}
Keiron O'Shea and Ryan Nash.
\newblock An introduction to convolutional neural networks, 2015.
\newblock URL \url{http://arxiv.org/abs/1511.08458}.

\bibitem[Pan et~al.(2023)Pan, Wang, Qiu, Axente, Chang, Peng, Patel, Shelton,
  Patel, Roper, and Yang]{pan_2d_2023}
Shaoyan Pan, Tonghe Wang, Richard L~J Qiu, Marian Axente, Chih-Wei Chang, Junbo
  Peng, Ashish~B Patel, Joseph Shelton, Sagar~A Patel, Justin Roper, and
  Xiaofeng Yang.
\newblock 2d medical image synthesis using transformer-based denoising
  diffusion probabilistic model.
\newblock \emph{Physics in Medicine \& Biology}, 68\penalty0 (10):\penalty0
  105004, 2023.
\newblock \doi{10.1088/1361-6560/acca5c}.
\newblock URL \url{https://dx.doi.org/10.1088/1361-6560/acca5c}.

\bibitem[Phan et~al.(2023)Phan, Liao, Verjans, and
  To]{phan_structure_preserving_2023}
Vu~Minh~Hieu Phan, Zhibin Liao, Johan~W Verjans, and Minh-Son To.
\newblock Structure-preserving synthesis: {MaskGAN} for unpaired mr-ct
  translation.
\newblock In \emph{International Conference on Medical Image Computing and
  Computer-Assisted Intervention}, pages 56--65. Springer, 2023.

\bibitem[Rombach et~al.(2021)Rombach, Blattmann, Lorenz, Esser, and
  Ommer]{rombach_high-resolution_2022}
Robin Rombach, Andreas Blattmann, Dominik Lorenz, Patrick Esser, and Björn
  Ommer.
\newblock High-resolution image synthesis with latent diffusion models.
\newblock \emph{CoRR}, abs/2112.10752, 2021.

\bibitem[Shamshad et~al.(2023)Shamshad, Khan, Zamir, Khan, Hayat, Khan, and
  Fu]{shamshad_transformers_2023}
Fahad Shamshad, Salman Khan, Syed~Waqas Zamir, Muhammad~Haris Khan, Munawar
  Hayat, Fahad~Shahbaz Khan, and Huazhu Fu.
\newblock Transformers in medical imaging: A survey.
\newblock \emph{Medical Image Analysis}, 88:\penalty0 102802, 2023.
\newblock ISSN 1361-8415.
\newblock \doi{10.1016/j.media.2023.102802}.
\newblock URL
  \url{https://www.sciencedirect.com/science/ARTICLE/pii/S1361841523000634}.

\bibitem[Sohl-Dickstein et~al.(2015)Sohl-Dickstein, Weiss, Maheswaranathan, and
  Ganguli]{dickstein_deep_2015}
Jascha Sohl-Dickstein, Eric Weiss, Niru Maheswaranathan, and Surya Ganguli.
\newblock Deep unsupervised learning using nonequilibrium thermodynamics.
\newblock In \emph{Int Conf Mach Learn}, pages 2256--2265, 2015.

\bibitem[Sohn et~al.(2015)Sohn, Yan, and Lee]{sohn_learning_nodate}
Kihyuk Sohn, Xinchen Yan, and Honglak Lee.
\newblock Learning structured output representation using deep conditional
  generative models.
\newblock In \emph{Proceedings of the 29th International Conference on Neural
  Information Processing Systems - Volume 2}, NIPS'15, page 3483–3491. MIT
  Press, 2015.

\bibitem[{Subtle Medical}(2019)]{subtle_medical_2019}
{Subtle Medical}.
\newblock Deep learning boosts healthcare. faster, safer, smarter.
\newblock \url{https://subtlemedical.com/subtlemr/}, 2019.

\bibitem[Vaswani et~al.(2017)Vaswani, Shazeer, Parmar, Uszkoreit, Jones, Gomez,
  Kaiser, and Polosukhin]{vaswani_attention_2023}
Ashish Vaswani, Noam~M. Shazeer, Niki Parmar, Jakob Uszkoreit, Llion Jones,
  Aidan~N. Gomez, Lukasz Kaiser, and Illia Polosukhin.
\newblock Attention is all you need.
\newblock \emph{Adv Neural Inf Process Syst}, pages 1--11, 2017.

\bibitem[Wang et~al.(2024)Wang, Heimann, Tannast, and
  Zheng]{wang_cyclesgan_2024}
Runze Wang, Alexander~F. Heimann, Moritz Tannast, and Guoyan Zheng.
\newblock {CycleSGAN}: A cycle-consistent and semantics-preserving generative
  adversarial network for unpaired {MR}-to-{CT} image synthesis.
\newblock \emph{Computerized Medical Imaging and Graphics}, 117:\penalty0
  102431, 2024.
\newblock ISSN 0895-6111.
\newblock \doi{10.1016/j.compmedimag.2024.102431}.
\newblock URL
  \url{https://www.sciencedirect.com/science/ARTICLE/pii/S0895611124001083}.

\bibitem[Wang et~al.(2025)Wang, Xia, Luo, Dong, Li, Wen, and
  Li]{wang_diffusion_2025}
Wei Wang, Jiayu Xia, Gongning Luo, Suyu Dong, Xiangyu Li, Jie Wen, and Shuo Li.
\newblock Diffusion model for medical image denoising, reconstruction and
  translation.
\newblock \emph{Computerized Medical Imaging and Graphics}, 124:\penalty0
  102593, 2025.
\newblock ISSN 0895-6111.
\newblock \doi{10.1016/j.compmedimag.2025.102593}.
\newblock URL
  \url{https://www.sciencedirect.com/science/ARTICLE/pii/S0895611125001028}.

\bibitem[Xiao et~al.(2021)Xiao, Kreis, and Vahdat]{xiao_tackling_2022}
Zhisheng Xiao, Karsten Kreis, and Arash Vahdat.
\newblock Tackling the generative learning trilemma with denoising diffusion
  {GANs}.
\newblock \emph{arXiv:2112.07804}, 2021.

\bibitem[Xing et~al.(2024)Xing, Yang, Chen, Ye, Yang, Qin, and
  Zhu]{linguraru_cross_conditioned_2024}
Zhaohu Xing, Sicheng Yang, Sixiang Chen, Tian Ye, Yijun Yang, Jing Qin, and Lei
  Zhu.
\newblock Cross-conditioned diffusion model for medical image to image
  translation.
\newblock In \emph{proceedings of Medical Image Computing and Computer Assisted
  Intervention -- MICCAI 2024}, volume LNCS 15007. Springer Nature Switzerland,
  October 2024.

\bibitem[Zeineldin et~al.(2023)Zeineldin, Karar, Burgert, and
  Mathis-Ullrich]{bakas}
Ramy~A. Zeineldin, Mohamed~E. Karar, Oliver Burgert, and Franziska
  Mathis-Ullrich.
\newblock Multimodal {CNN} networks for brain tumor segmentation in {MRI}: A
  {BraTS} 2022 challenge solution.
\newblock In Spyridon Bakas, Alessandro Crimi, Ujjwal Baid, Sylwia Malec,
  Monika Pytlarz, Bhakti Baheti, Maximilian Zenk, and Reuben Dorent, editors,
  \emph{Brainlesion: Glioma, Multiple Sclerosis, Stroke and Traumatic Brain
  Injuries}, volume 13769, pages 127--137. Springer Nature Switzerland, 2023.
\newblock URL \url{https://link.springer.com/10.1007/978-3-031-33842-7_11}.

\bibitem[Zhu et~al.(2017)Zhu, Park, Isola, and Efros]{zhu_unpaired_2020}
Jun-Yan Zhu, Taesung Park, Phillip Isola, and Alexei~A. Efros.
\newblock Unpaired image-to-image translation using cycle-consistent
  adversarial networks.
\newblock In \emph{2017 IEEE International Conference on Computer Vision
  (ICCV)}, pages 2242--2251, 2017.
\newblock \doi{10.1109/ICCV.2017.244}.

\bibitem[Zhu et~al.(2024)Zhu, Zhang, Li, O'Donnell, and Zhang]{zhu_2024}
Xi~Zhu, Wei Zhang, Yijie Li, Lauren~J. O'Donnell, and Fan Zhang.
\newblock When diffusion {MRI} meets diffusion model: A novel deep generative
  model for diffusion {MRI} generation, 2024.

\bibitem[Özbey et~al.(2023)Özbey, Dalmaz, Dar, Bedel, Özturk, Güngör, and
  Çukur]{ozbey_unsupervised_2023}
Muzaffer Özbey, Onat Dalmaz, Salman U.~H. Dar, Hasan~A. Bedel, Şaban Özturk,
  Alper Güngör, and Tolga Çukur.
\newblock Unsupervised medical image translation with adversarial diffusion
  models.
\newblock \emph{IEEE Transactions on Medical Imaging}, 42\penalty0
  (12):\penalty0 3524--3539, 2023.
\newblock \doi{10.1109/TMI.2023.3290149}.

\end{thebibliography}

\end{document}